\documentclass[letterpaper, 10 pt, conference]{ieeeconf}  
\IEEEoverridecommandlockouts                             
                                                         
\usepackage{graphicx} 
\usepackage{amsmath} 
\usepackage{amssymb}  
\usepackage[tight,footnotesize]{subfigure}
\usepackage{soul}
\usepackage{color}
\usepackage{multirow}
\usepackage{pbox}
\usepackage{flushend}
\usepackage{pdflscape} 
\usepackage{wasysym}
\usepackage[normalem]{ulem}

\definecolor{CommentRed}{rgb}{0.7,0,0}
\definecolor{CommentGreen}{rgb}{0,0.7,0}
\definecolor{CommentBlue}{rgb}{0,0,0.7}

\newcommand{\graveyard}[1]{}
\newcommand{\state}{y}
\newcommand{\hidden}{h}

\newcommand{\observation}{x}

\title{\LARGE \bf
Deep Tracking on the Move: Learning to Track the World from a Moving Vehicle using Recurrent Neural Networks
}

\author{Julie Dequaire, Dushyant Rao, Peter Ondr\'{u}\v{s}ka, Dominic Zeng Wang and Ingmar Posner%
\thanks{Authors are from the Mobile Robotics Group at the University of Oxford, United Kingdom: 
     {\tt\small julie, dushyant, ingmar@robots.ox.ac.uk}}%
        }

\begin{document}
\maketitle
\thispagestyle{empty}
\pagestyle{empty}

\begin{abstract}
This paper presents an end-to-end approach for tracking static and dynamic objects for an autonomous vehicle driving through crowded urban environments.
Unlike traditional approaches to tracking, this method is learned end-to-end, and is able to directly predict a full unoccluded occupancy grid map from raw laser input data.
Inspired by the recently presented \emph{DeepTracking} approach (\cite{OndruskaAAAI2016},~\cite{ondruska2016end}), we employ a recurrent neural network (RNN) to capture the temporal evolution of the state of the environment, and propose to use \emph{Spatial Transformer} modules to exploit estimates of the egomotion of the vehicle.
Our results demonstrate the ability to track a range of objects, including cars, buses, pedestrians, and cyclists through occlusion, from both moving and stationary platforms, using a single learned model.
Experimental results demonstrate that the model can also predict the future states of objects from current inputs, with greater accuracy than previous work.


\end{abstract}

\section{Introduction}
The safe and effective operation of an autonomous vehicle depends on its ability to interpret its surroundings and track and predict the state of the environment over time.
Many tracking systems employ multiple hand-engineered stages (e.g. object detection, semantic classification, data association, state estimation and motion modelling, occupancy grid generation) in order to represent the state and evolution of the world (\cite{Vu2007,Wang01062015,Petrovskaya2009}). However, as the tasks assigned to robots become more complex, this approach becomes increasingly infeasible.

Recent advances in machine learning, particularly those of deep neural networks, have demonstrated the ability to capture complex structure in the real world, and have led to significant improvements in numerous computer vision and natural language processing applications (\cite{krizhevsky2012imagenet, dahl2012context, wang2012end}). Such approaches would however typically require large, task-specific corpora of annotated ground-truth labels to master the desired task. This becomes difficult when learning a model of the environment without access to corresponding ground truth, as is often the case for object tracking in crowded urban environments.

In recent work,~\cite{ondruska2016end} took an alternative approach and presented an end-to-end fully and efficiently trainable framework for learning a model of the world dynamics, building on the original \emph{DeepTracking} work by~\cite{OndruskaAAAI2016}. We considered the specific problem of learning to track and classify moving objects in a complex and only partially-observable real-world scenario, as viewed from a \emph{static} sensor. Here, we advance this work and address the problem of tracking from a \emph{moving} platform. We extend the neural network architecture proposed in~\cite{ondruska2016end} to account for the egomotion of the sensor frame as it moves in the world frame, and demonstrate improved tracking accuracy as compared to previous work.

We demonstrate the system on laser point cloud data collected in a busy urban environment, with an array of static and dynamic objects, including cars, buses, pedestrians and cyclists.
The model not only bypasses the sequence of hand-engineered steps typical of traditional tracking approaches, but is empirically shown to successfully predict the future evolution of objects in the environment, even when they are completely occluded.

\begin{figure}[t]
\centering
\includegraphics[width=8.4cm]{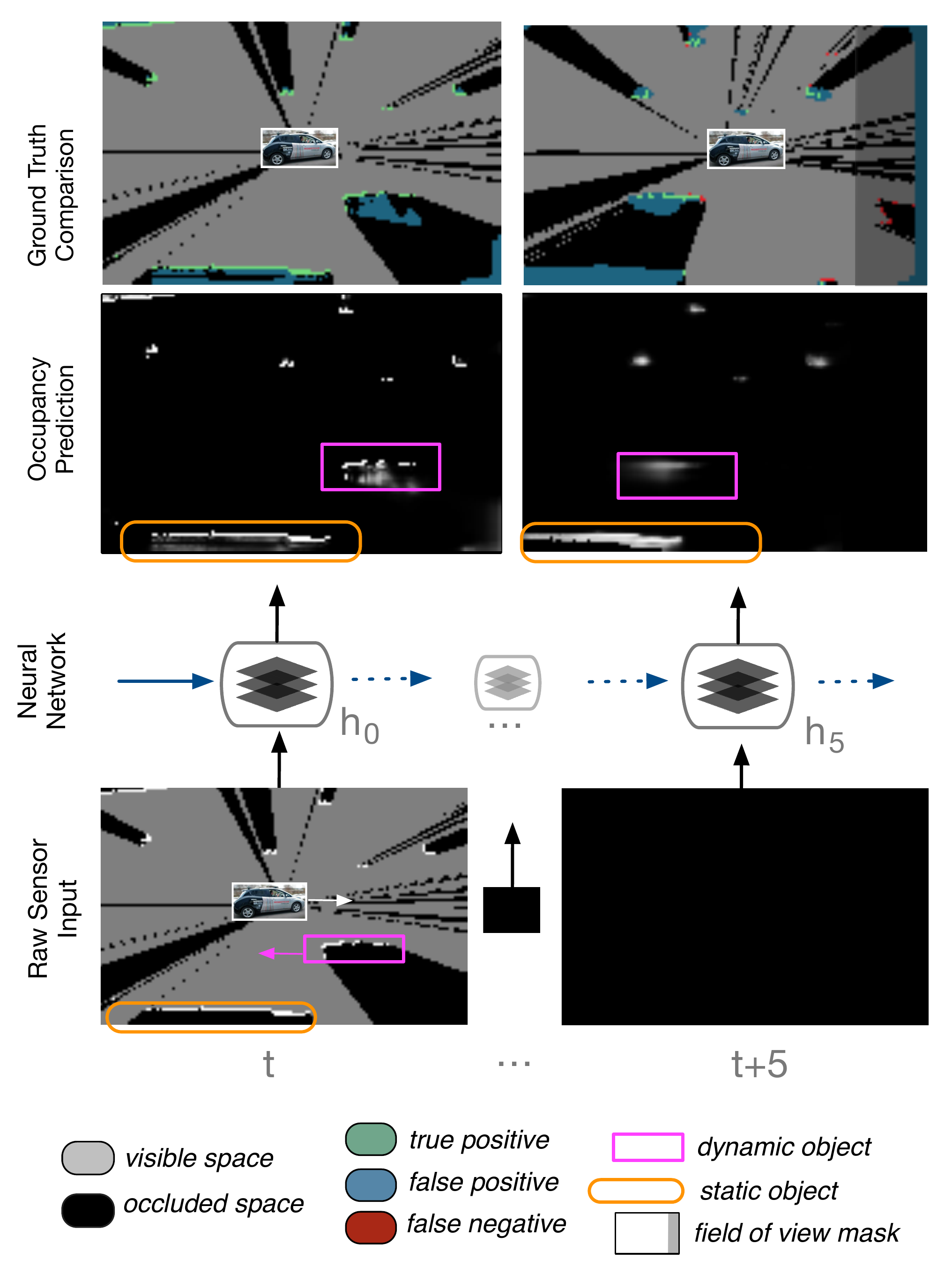}
\caption{A typical training sequence. The unoccluded occupancy map is predicted directly from the input grid data, allowing objects to be tracked in occlusion and in future frames. The observed false positives are therefore beneficial. Comparison to the visible ground truth shows that the model is able to capture the dynamics of the moving vehicle (pink rectangle) and accurately predicts its track.}
\label{fig:dt_arch}
\vspace{-0.4cm}
\end{figure}

The rest of the paper is structured as follows. Section~\ref{sec:related} highlights related work and Section~\ref{sec:formulation} summarises the problem definition and \emph{DeepTracking} framework first presented in~(\cite{OndruskaAAAI2016, ondruska2016end}). Section~\ref{sec:overview} describes the models used to perform tracking in real-world scenarios considering both static and dynamic sensors. 
Section~\ref{sec:results} presents an empirical evaluation of our methods, and Section~\ref{sec:conclusion} concludes the paper and discusses the future implications of our findings.

\section{Related Work}
\label{sec:related}
A number of previous works have explored the problem of model-free tracking of objects in the environment of an autonomous vehicle (\cite{ Vu2007,Wang01062015,Petrovskaya2009}).
Typically, these approaches follow the traditional paradigm of a multi-component pipeline, with separate components to parametrise and detect objects, associate new measurements to existing tracks, and estimate the state of each individually tracked object.
The use of multiple stages in the framework is cumbersome and introduces extra unnecessary failure modes for the tracking algorithm.

Recent work proposes to replace these multiple stages with an end-to-end learning framework known as \emph{DeepTracking}, by leveraging neural networks to directly learn the mapping from raw laser input to an unoccluded occupancy grid~(\cite{OndruskaAAAI2016, ondruska2016end}), even with relatively small amounts of data.
The approach utilises an RNN architecture using \textit{gated recurrent units}~\cite{cho2014learning} to capture the state and evolution of the world in a sequence of laser scan frames. Another work \cite{choirobust} considers \emph{recurrent flow networks} and takes a different angle to predicting occupancy in dynamic environments. They explicitly encode a range of velocities in the hidden layers of a recurrent network, and use Bayesian optimization to learn the network parameters which update velocity estimation and occupancy prediction. However, the model does not explicitly track objects through occlusion.

\emph{DeepTracking} shares similarities with deep learning approaches to predictive video modelling (\cite{patraucean2015spatio, lotter2016deep}), in that it is trained to predict the future state of the world based on current input data.
This is particularly important, because in order to successfully capture the future location of dynamic objects in the scene, the model must implicitly store the position and velocity of each object in its internal memory state.

While this eliminates the need to design individual components by hand, the model assumes a static vehicle~\cite{ondruska2016end}.
Extending the problem to a moving platform is a challenging task, as it introduces an array of complex relative motions between the vehicle and objects in its environment.
As \emph{DeepTracking} ignores the motion of the vehicle, the model is forced to learn all possible motion interactions between the vehicle and its environment as if the vehicle were stationary. For a moving platform, we leverage estimates of egomotion as a proxy for vehicle motion.
We scale up the RNN-based models proposed by~(\cite{OndruskaAAAI2016, ondruska2016end}) for real-world application on dynamic vehicles, and exploit \emph{Spatial Transformer} modules \cite{jaderberg2015spatial}, which allow the internal memory state representations to be spatially transformed according to the estimated egomotion.
\newpage
The main contributions of this work are as follows:
\begin{enumerate}
    \item A more in-depth analysis of the performance of \emph{DeepTracking} in the case of a static vehicle, building on the experiments presented in \cite{ondruska2016end}.
    \item The use of \emph{Spatial Transformer} modules to exploit estimates of visual egomotion in the \emph{DeepTracking} framework.
    \item Demonstration of end-to-end tracking of a variety of object classes through occlusion, on a moving vehicle in crowded urban scenes.
\end{enumerate}

\section{Tracking Problem Formulation}
\label{sec:formulation}
The problem we address in this paper is to uncover the true state of the world, in terms of a 2D occupancy grid $y_t$, given a sequence of \emph{partially observed} states of the environment $x_{1:t}$ computed from raw sensor measurements. 
In particular, we solve for $P(y_t|x_{1:t})$, the probability of the true unoccluded state of the world at time $t$ given a sequence of partial observations at all previous time steps. This formulation can also be used to predict future states by solving for $P(\state_{t+n}|\observation_{1:t})$, given empty input for $\observation_{t+1:t+n}$.

\begin{equation}
P(\state_{t}|\observation_{1:t}) = P(\state_t | \hidden_t )
\label{eq:prediction}
\end{equation}

Evolution of this latent state $\hidden_t$, which includes propagating model dynamics and integrating new sensor measurements, is modelled by the update equation:
\begin{equation}
\hidden_{t} = f(\hidden_{t-1},\observation_{t})
\label{eq:update}
\end{equation}

The key element here is that both the latent state update $f(\hidden_{t-1},\observation_{t})$, and the decoding step to produce the output $P(\state_t | \hidden_t )$ are modelled as parts of a single neural network and are trained jointly. Equations \ref{eq:prediction} and \ref{eq:update} can then be performed repeatedly as the building block of a recurrent neural network that continuously updates the belief state $h_t$, and uses it as network memory to predict $y_t$. This makes it suitable for online stream filtering of sensor input.

When the output ground-truth $y_t$ is not readily available, as is the case in real-world scenarios, the network can be trained in an self-supervised fashion. This is made possible by considering that predicting the movement of objects in occlusion at time $t$ is similar to predicting a \emph{future} state $y_{t+n}$ provided no input is given to the network, i.e $x_{t+n}=\diameter$. Lack of input observation equates to complete occlusion of the scene. As only \emph{observable} ground truth is available, we reduce the problem of predicting $y_{t+n}$ to that of predicting the directly observable input $x_{t+n}$. Training the network to predict $P(x_{t+n}|x_{1:t})$ is then equivalent to computing and backpropagating the errors only on the observable parts of the scene.
We refer the reader to~\cite{OndruskaAAAI2016} and~\cite{ondruska2016end} for further details on the RNN and the training procedure.

Each input observation $\observation_{t}~\in~\{0,1\}^{2 \times M \times M}$ is represented as a pair of discretised 2D binary grids of size $M\times M$, parallel to the ground and locally built around the sensor. The first matrix encodes whether a cell is directly observable by the sensor at time $t$, while the second matrix encodes whether a cell is observed to be free (value of $0$) or occupied (value of $1$). We refer to these two matrices as $x_{t,vis}$ and $x_{t,occ}$, the \emph{visibility} and \emph{occupancy} grids respectively. The output we wish to obtain is an occlusion-free state of the world $y_t~\in~\{0,1\}^{M \times M}$, and is represented by an occupancy matrix similar to that of the input occupancy grid of $x_t$.    

In the next section, we build upon~\cite{ondruska2016end} to deploy the \emph{DeepTracking} paradigm on a real-world moving platform.

\section{Technical Overview}
\label{sec:overview}
\subsection{Deep Tracking from a Static Sensor}
\label{sec:DTstatic}
First, we revisit~\cite{ondruska2016end} to detail the baseline \emph{DeepTracking} architecture for real world application, which we extend to a moving platform.

At each time step $t$, the partially observed grid $\observation_{t}$ used as input to the network is computed from raw 2D laser scans by ray tracing. Cells in which a measurement ends are marked as occupied, all cells from the sensor origin up to the end of the ray are marked as free, and cells beyond the ray are marked to be unobserved.

\begin{figure}[t]
    \centering
    \includegraphics[width=90mm,trim={0 1cm -1.cm 0},clip=true]{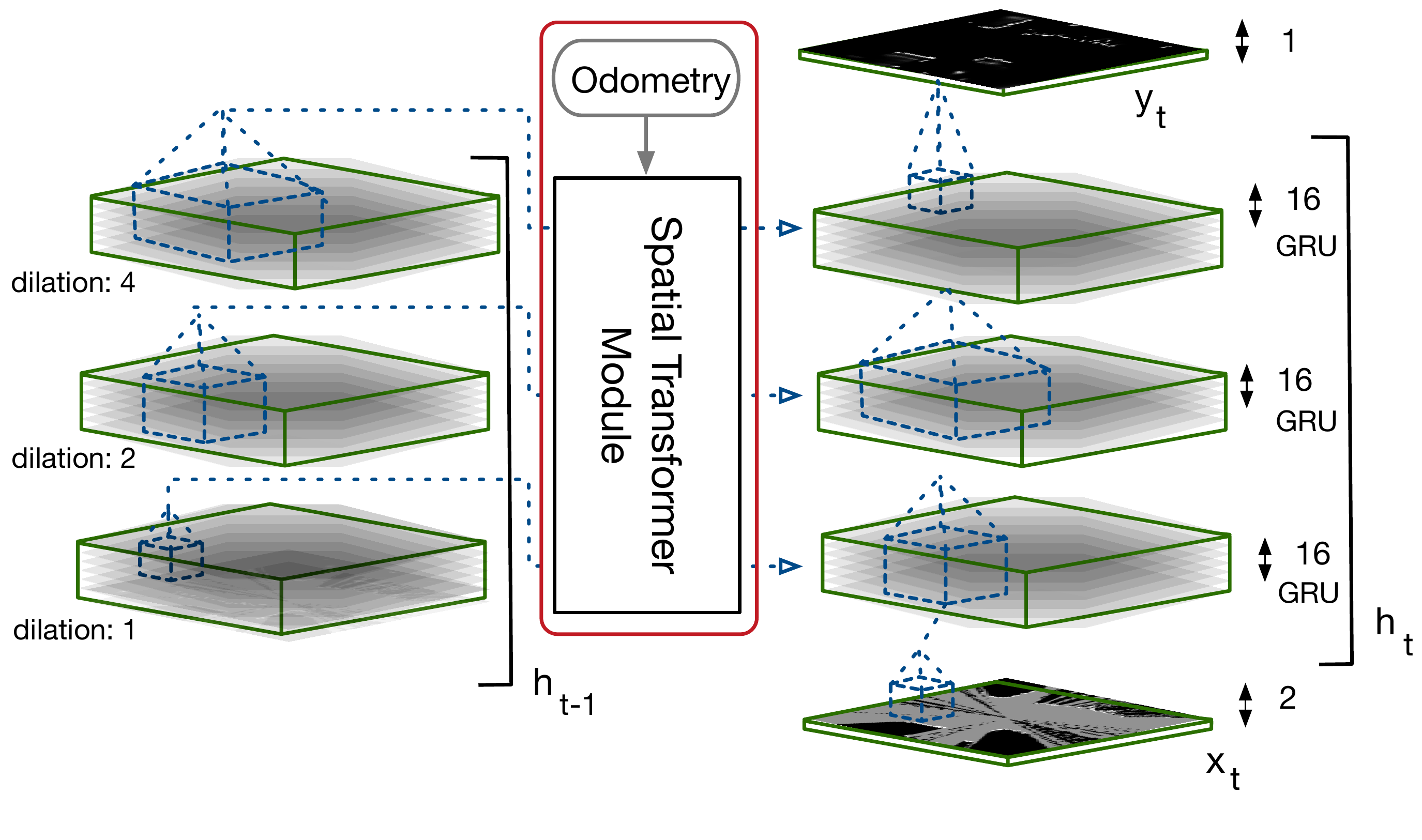}
    \caption{The proposed network architecture features dilated convolutions, gated recurrent units, a spatial transformer module, and outputs cell occupancy in the sensor's surroundings. The spatial transformer module is only utilised in a moving vehicle scenario.}
    \label{fig:architecture}
\end{figure}

The input $\observation_{t}$ is processed by a multi-layer network illustrated in Figure~\ref{fig:architecture}, with the Spatial Transformer module only utilised in the moving vehicle scenario.
The architecture originally proposed in~\cite{OndruskaAAAI2016} is scaled up with dilated convolutions~\cite{yu2015multi} and a variant of gated recurrent units (\cite{cho2014learning, xingjian2015convolutional}) to allow for the simultaneous tracking of different-sized objects such as cars and pedestrians.
Each layer $l$ at time $t$ is updated considering its own activations at time $t-1$ and those of layer $l-1$ below at time $t$, thus implementing the recurrence.
This allows the network to extract and remember the information from the past and to use it for prediction of occluded objects or future states of the world. An additional static memory can also be utilised, in which the network is able to learn individual pixel-wise biases to add to the output of every convolutional operation.
The output of the final layer is then converted into the output cell occupancy $\state_{t}$ via a simple convolutional decoder. As explicited in the results section, we also experiment with architectures that decode the entire hidden state to the output.

\subsection{Deep Tracking from a Moving Vehicle}    
\label{sec:DTdyn}

Tracking a dynamic scene from a moving platform poses the challenge of decoupling the motion of the vehicle from the motion of objects in the global environment. 
In the static scenario, information related to an object located at index $\{i,j\}$ in the input $\observation_{t}$ would typically be stored at the corresponding neighbouring spatial location $\{i+\Delta i, j+\Delta j\}$ in the layers of the latent representation, with the neighbourhood $\Delta_{i,j}$ based on the receptive field of each neuron in the hidden state. The latent state update would then pass the information along spatially in accordance with the observed relative motion of the input object (Figure~\ref{fig:hiddStatic}).
In the static scenario, the dynamics of the scene as viewed from the world frame are coherent with that viewed from the local sensor frame.

When tracking from a moving vehicle however, the spatial update of information within the latent representation would additionally need to account for the sensor's egomotion as it will affect the position of the object relative to the sensor frame.
Although this is a major drawback of the baseline \emph{DeepTracking} architecture, it can be compensated by transforming the memory state in accordance to the egomotion. In other words, a static obstacle situated at position $\{i_{t-1},j_{t-1}\}$ in the sensor frame at $t-1$, will be moved to position $\{i_{t},j_{t}\}$ in the sensor frame at time $t$ such that:

\begin{equation}
 [x_{t},y_{t},1]^{T}= T_{t,t-1}\times[x_{t-1},y_{t-1},1]^{T}
\label{eq:transform}
\end{equation}

\noindent where $T_{t,t-1}$ is the $\mathbb{SE}(2)$ forward transformation of the sensor source frame at $t-1$ into the sensor destination frame at $t$. This formulation naturally extends to 3D motion with $\mathbb{SE}(3)$.

We aid the network in decoupling egomotion and object motion in this way by introducing a \emph{Spatial Transformer}~\cite{jaderberg2015spatial} module (STM) in the hidden state (Figure~\ref{fig:architecture}). In the original work by~\cite{jaderberg2015spatial}, the STM is introduced as a learnable module for actively spatially transforming feature maps to aid tasks such as the classification of distorted datasets.
However, in the context of tracking from a moving platform where egomotion is readily available (e.g. visual odometry), the STM can be used just to forward transform the hidden feature maps centred in the sensor source frame at time $t-1$, into the sensor destination frame at time $t$, using transformation $T_{t,t-1}$ (Equation~\ref{eq:transform}).

Thus, the STM is introduced in the hidden state and performs a transformation on all feature maps of $h_{t-1}$, set in frame $t-1$, into frame $t$ before update with the new incoming input $x_{t}$ to form the new memory at $t$.

\subsection{Training}
In both static and dynamic cases, the network is presented the input sequence $x_{1:t}$ and trained to predict the next $n$ input frames $x_{t+1:t+n}$.
The binary cross-entropy loss is calculated and backpropagated only on the visible part of the output, which is achieved by simply masking the prediction $y_{t}$ with $x_{t,vis}$ and multiplying the resulting grid element-wise with the occupancy part grid $x_{t,occ}$.
By using a loss that encourages the model to predict the state of the environment in the future, the model is forced to capture the motion of each object in the hidden representation.

In the static sensor scenario, the error is only backpropagated on the ground truth available, i.e the \emph{visible} part of the space.
In the case of a moving sensor, however, an additional constraint needs to be imposed to account for the fact that as the robot moves in future frames, it will discover new space that falls outside the field of view of the current frame.
Given this fact, the model should not be falsely penalised for failing to accurately guess objects located within this new space when the input is blanked out.
This is similar in nature to the static case, where the input grid also represents a frontier between what the robot can perceive and understand of the scene and the unknown world outside its field of view.

To address this, we apply an additional mask at training time on the cost computation and error backpropagation to represent the \emph{predictable} space in future frames. Accounting for this field of view drift is crucial in terms of tracking performance, as it corrects for an objective function that is otherwise skewed towards the incredibly difficult task of predicting objects outside the field of view. This mask has been overlaid in transparency over the ground truth comparison outputs of Figure \ref{fig:dt_arch}, and indicates the predictable free space shrinking on future timesteps. 

\section{Experimental Results}
\label{sec:results}
In this section, we perform experimental validation of both the baseline \emph{DeepTracking} and the STM-based variant proposed in this paper.
With a stationary vehicle, no spatial transform is necessary and the two are identical.

\subsection{Static Vehicle}
For the static case, we consider an architecture with three hidden layers of 16 GRU feature maps each. Computation of the hidden state consists of $3\times3$ convolutional filters, applied on the three layers (from bottom up) with strides of 1, 2, and 4 pixels, and receptive fields of $3\times3$, $7\times7$, and $15\times15$, respectively. We consider an input field of view of $20\times20$ $m^{2}$ discretised into cells of $20\times20$ $cm^{2}$, which translates into a $H\times W=101\times101$ input grid. With a hidden state consisting of 48 feature maps, the additional static memory contributes to $H\times W \times D=489,648$ of the total $1,506,865$ parameters of the network.

The evaluation dataset consists of a $75$ minute log collected from a stationary \emph{Hokuyo} UMT-30LX 2D laser scanner, positioned in the middle of a busy urban intersection. The area features dense traffic composed of buses, cars, cyclists and pedestrians, which results in extensive amounts of occlusion. Consequently, at no point in time is the complete scene fully observable. The dataset is subsampled at $8$Hz and split into a $65$ minute set for training, and a $10$ minute set for testing occupancy prediction.

The full model was trained on an Nvidia Tesla  K40 GPU until convergence, using the unsupervised training procedure described in Section~\ref{sec:formulation}.
The training set is split into mini-batch sequences of 40 frames (5 seconds).
For every mini-batch, the network is shown 10 frames and trained to predict the next 10 frames, leading to two such sequences per 40-frame mini-batch.
This length of sequence covers the typical lengths of occlusions observed but can be tuned accordingly.

\begin{figure}[t]
\centering
\includegraphics[width=8.4cm]{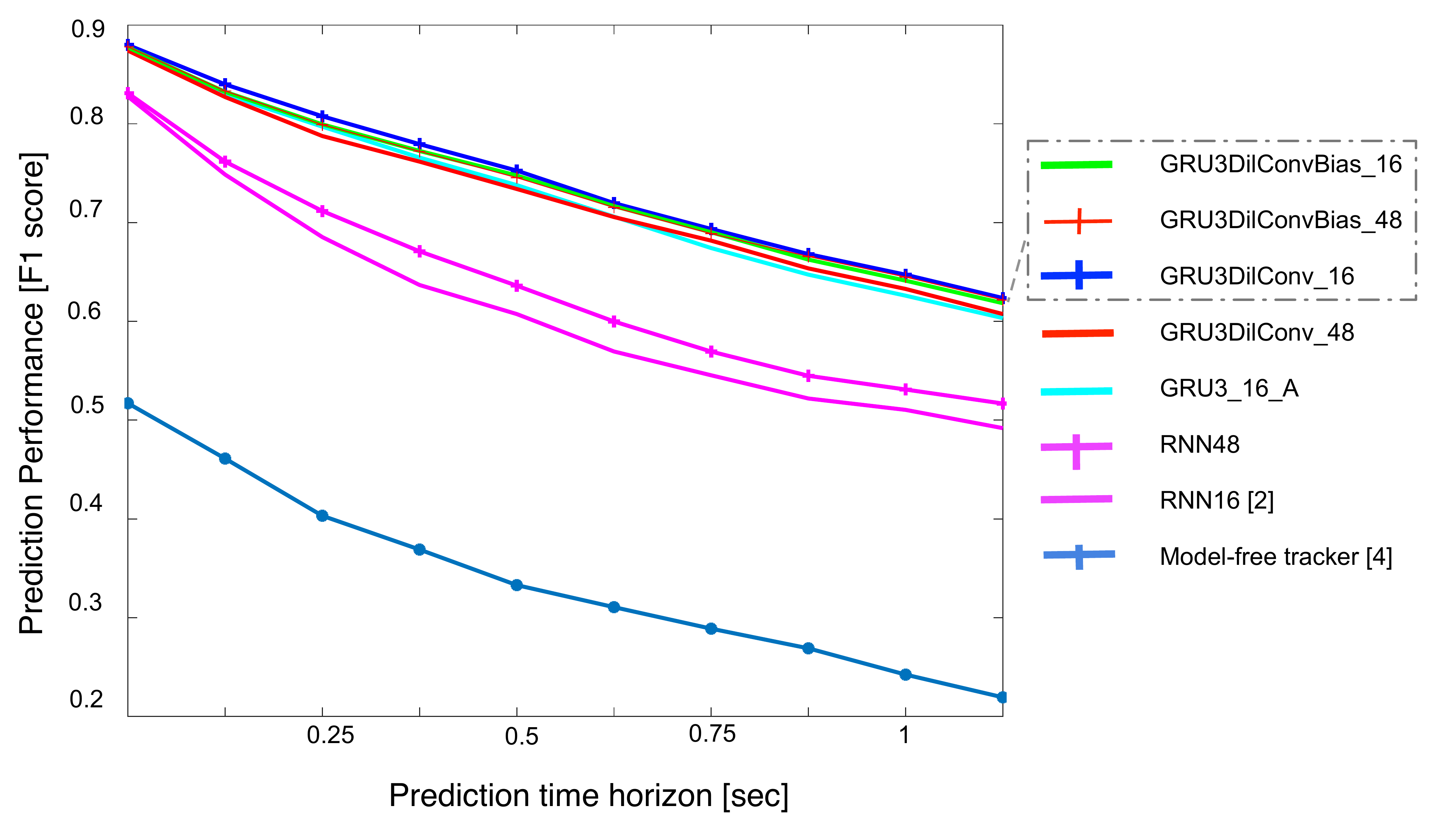}
\caption{F1 scores of network architectures when attempting to predict the future occupancy of the scene in a 1.25 second time horizon. The F1 measure is computed with a threshold of $0.5$ when considering a cell to be predicted as occupied or free.}
\label{fig:models}
\end{figure}

\begin{figure*}[t]
\centering
\includegraphics[width=17cm]{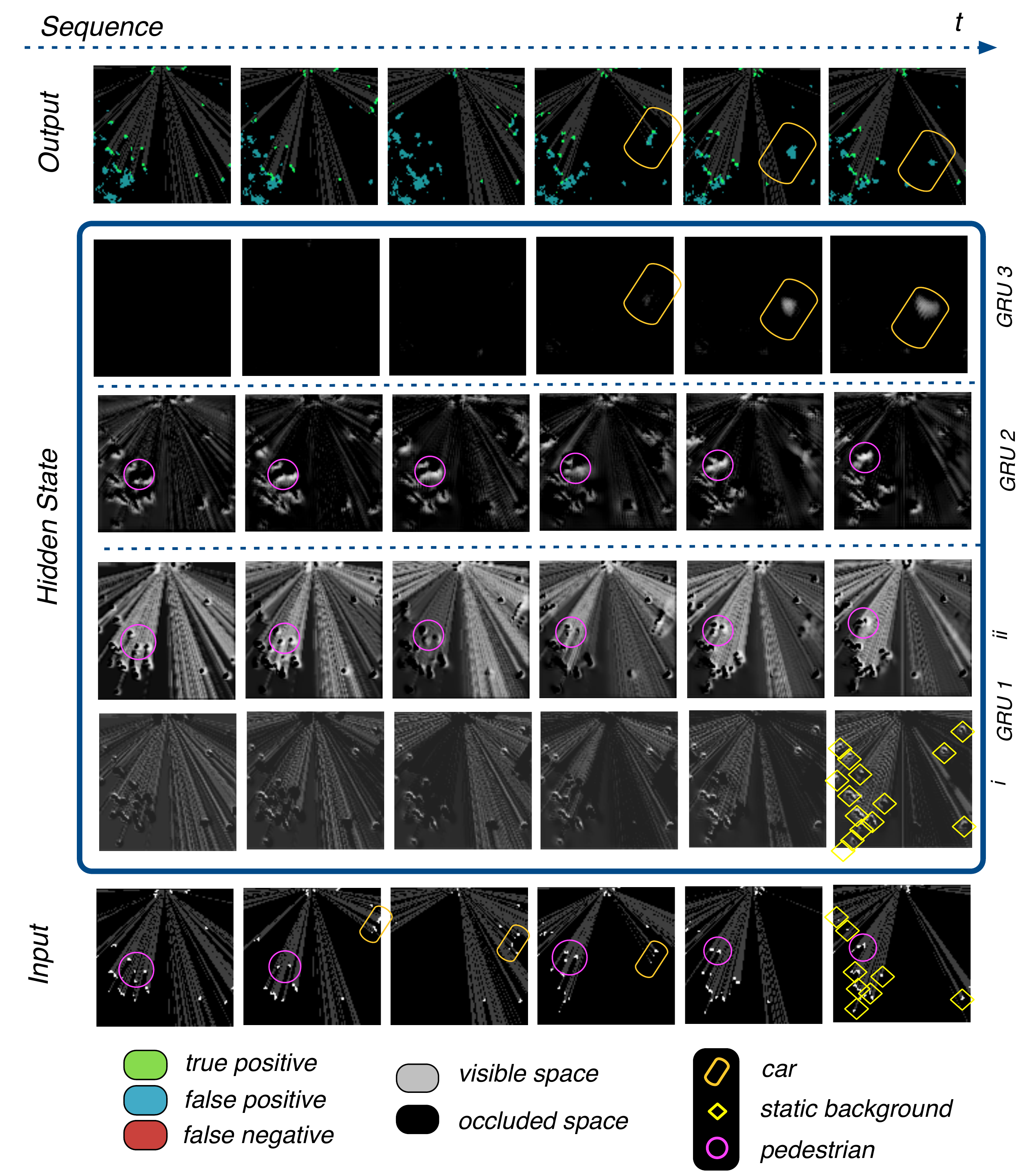}
\caption{Example of outputs produced by the system along with a selection of activations in the hidden layers. As highlighted in colour-coded circles, the network is able to propagate the assumed motion of the objects even when in complete occlusion. The sample hidden layer activations shown highlight the fact that lower layers in the hidden units (corresponding to a low dilation of the convolutions) capture the motion of small and slow moving objects such as pedestrians (e.g pink circle) and static background (e.g yellow squares), whereas a higher level layer learns to detect moving vehicles (orange oval).}
\label{fig:hiddStatic}
\end{figure*}

\subsubsection{Quantitative results}
We first look to quantify the gain in performance achieved by scaling up the original \emph{DeepTracking} network with the proposed architecture.
We compare a number of different architectures ranging from the original~\cite{OndruskaAAAI2016} to the proposed model in~\cite{ondruska2016end}, and compare performance on the task of predicting the observable near-future scene occupancy given the input sequence $x_{1:t}$. 
The predicted output occupancy $P(y_{t+n}|x_{1:t})$ is compared to $x_{t+n, occ}$, which corresponds to the observed occupancy of the world at time $t$.
A threshold of $0.5$ is used to determine whether a cell is predicted as occupied or free, and an F1 measure computed for each frame.

Figure~\ref{fig:models} compares the F1 measures computed on each of $n=10$ blacked out future frames, given the $10$ frames in the past.
All model predictions decrease over time as would be expected, as the uncertainty of the state of the world increases with the prediction horizon.
There is a notable step change in performance with neural architectures compared to a state-of-the art model-free tracking pipeline approach by~\cite{Wang01062015}. If we increase the capacity of the original RNN architecture of~\cite{OndruskaAAAI2016} (RNN16) from 16 to 48 feature maps (RNN48), we obtain marginal performance increase.
In comparison, replacing the standard RNN48 hidden unit state with three layers of 16 GRU units (GRU3$\_$16$\_$A) provides significant improvement in prediction ability. The affix \emph{16} signifies that we decode only the last hidden unit (composed of 16 features maps) to the output. Incorporating dilated convolutions (GRU3DilConv$\_$16) in place of traditional dense convolutions achieves comparable performance to GRU3DilConv$\_$16 for a similar output receptive field of 15$\times$15. The model with dilated convolutions additionally requires less computation and nearly one third fewer model parameters than its dense counterpart GRU3DilConv$\_$16. 
Further, we experiment with GRU3DilConv$\_$48 which decodes the \emph{full} hidden state (composed of 48 features maps) to the output. Performance of the model is maintained as illustrated in Figure~\ref{fig:models}. This is a departure from traditional architectures where information from the different scales of the hidden units is directly passed to the output. This may assist positively for tasks such as semantic labelling, where scale information is essential.
Lastly, performance of the full model with added static biases (GRU3DilConvBias$\_$16 and GRU3DilConvBias$\_$48) remains commensurate to that of GRU3DilConv$\_$16 and GRU3DilConv$\_$48 (Figure~\ref{fig:models}), and the learned static bias values may convey useful information such as that of the static background layout.
We consider the GRU3DilConv$\_$48 model in the rest of this paper.

\subsubsection{Qualitative results}
To better understand what the network has learned, we also qualitatively analyse a typical test sequence of length 3 seconds from  GRU3DilConv$\_$48, along with the network output and selected hidden state feature maps in Figure~\ref{fig:hiddStatic}.
The network is able to track pedestrians through full occlusion and the unobserved hallucinated tracks are represented in blue in the output sequence.

It can be seen that the GRU1 feature maps appear to have learned to capture the static background, as activations remain stationary during the sequence ($i$), and track pedestrians, as highlighted through the pink circles ($ii$).
The GRU2 feature map also captures the motion of pedestrians moving upwards to the left, while, interestingly, the GRU3 unit seems to activate only on the car that appears to the top right at frame 2 (indicated by the orange box).
This provides empirical support for the use of Dilated Convolutions which allow the model to capture patterns of increasing receptive fields in the hidden state units, while requiring fewer parameters than more traditional dense convolutional kernels.

In general, we observe that information regarding objects in the scene is captured in the hidden state, and moves spatially through the feature maps according to the motion of the object in the input. 
This can be problematic when extending tracking to a moving platform, as the object motion and vehicle motion are coupled.
We address this concern in the following section.

\subsection{Moving vehicle}
In this section we compare our baseline model GRU3DilConv$\_$48 (BaselineDT) with an equivalent architecture that incorporates the Spatial Transformer module into the hidden belief state (STM). As with the static case, we evaluate the ability to predict future frames given blacked out input, and illustrate the achieved occlusion-free tracking performance on a series of examples selected from the test set.

The evaluation dataset was collected over a 35 minute period, from a moving vehicle equipped with two Velodyne HDL64E lasers, resulting in a 360 degree field of view. The 3D point clouds were reduced to a 2D scan by considering the range of points within 0.6-1.5 meters height from the ground.

The network was trained on mini-batches of 40 sequences with a frame rate of $10$Hz, alternating between 5 inputs shown, and 5 inputs hidden. This higher frequency is better adapted to the moving vehicle case given the input field of view of $18\times18$ $\textrm{m}^{2}$ and a vehicle mean velocity of 20 miles per hour. Longer occluded sequences would lead to increased loss in useful memory due to the aforementioned drift of field of view.
Finally, we performed an $80/20$ split of the data into training and test set with no overlap in location, and trained our model on an Nvidia Tesla K40 GPU until convergence. Our architecture is implemented on Torch and uses the \emph{Spatial Transformer} GPU-implementation of~\cite{stnbhwd}.

\begin{figure}[t]
    \centering
    \includegraphics[width=71mm]{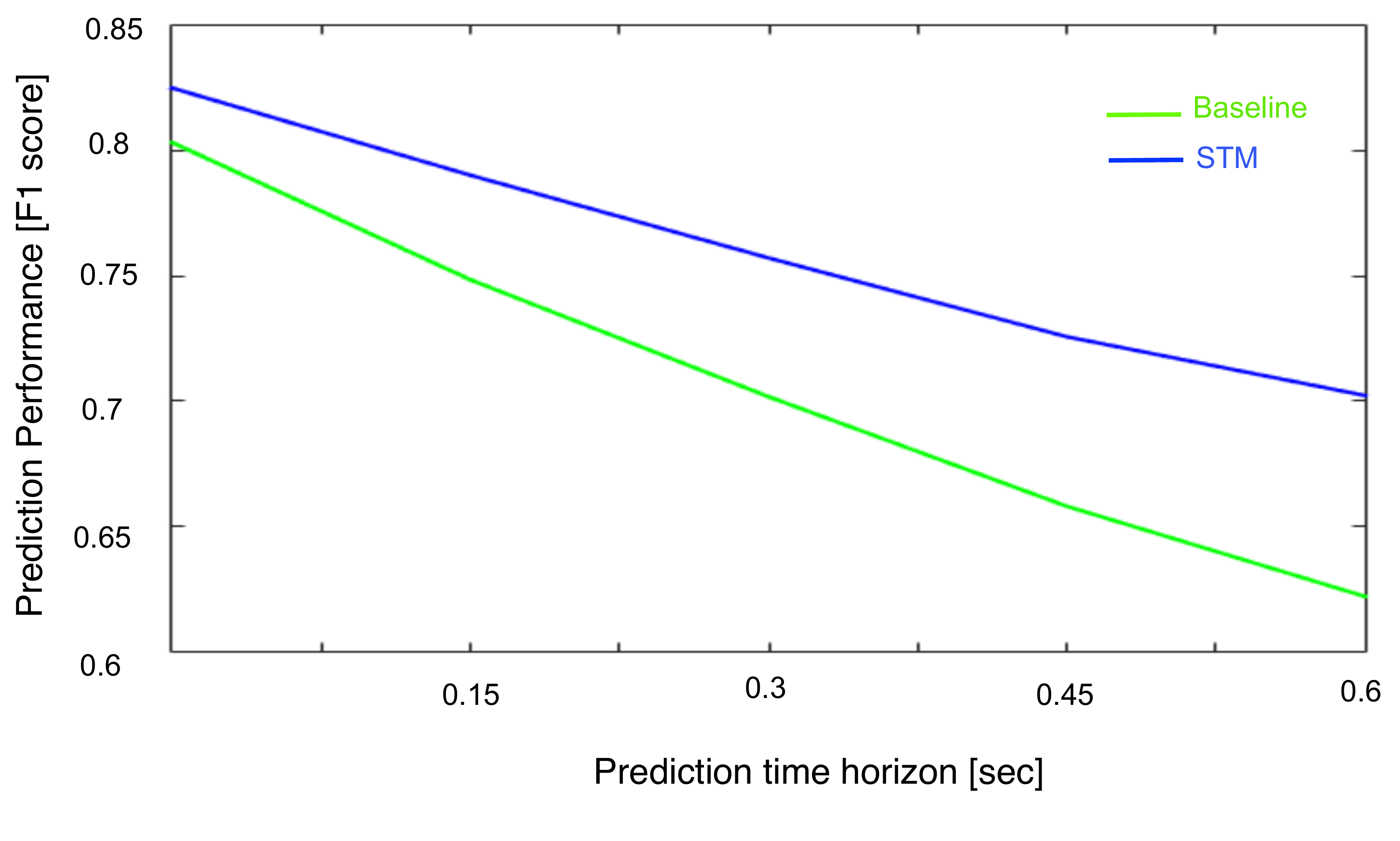}
    \caption{Positive contribution of the Spatial Transformer to the network's ability to correctly predict the future occupancy of the scene in a 0.6 sec time horizon. The baseline DT does surprisingly well which we attribute to the benign test set where the vehicle mostly evolves at constant velocity down straight roads. }
    \label{fig:F1moving}
\end{figure}

\begin{figure*}[t]
\centering
\includegraphics[width=18cm]{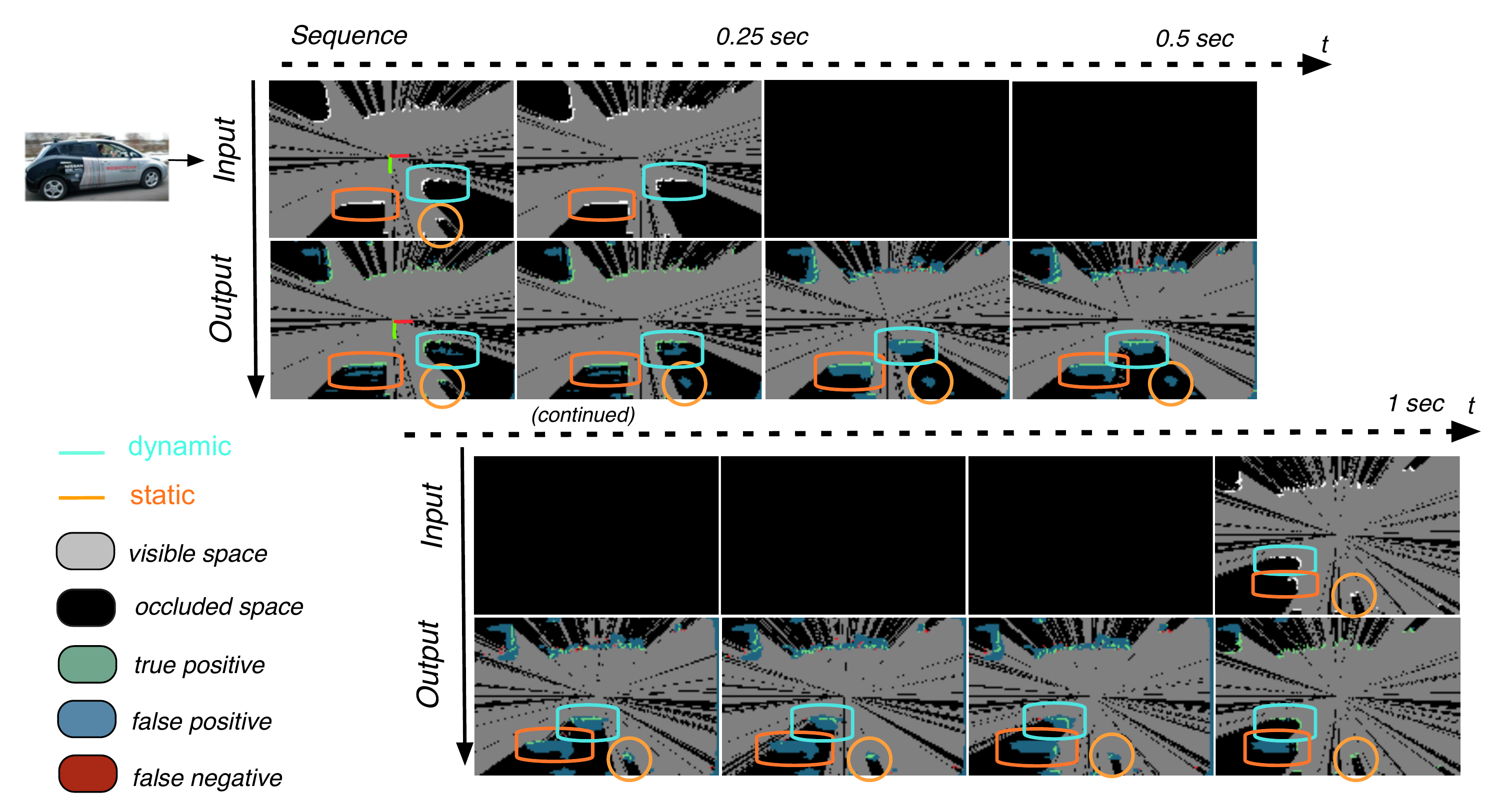}
\caption{Left to Right, Top to Bottom: Example sequence (1 second) of dynamic and static object tracking through occlusion by the STM. The output of the network is evaluated against the visible ground truth. We highlight several objects of interest that are correctly tracked through complete occlusion.}
\label{fig:ST1}
\end{figure*}

\begin{figure*}[t]
\centering
\includegraphics[width=18cm]{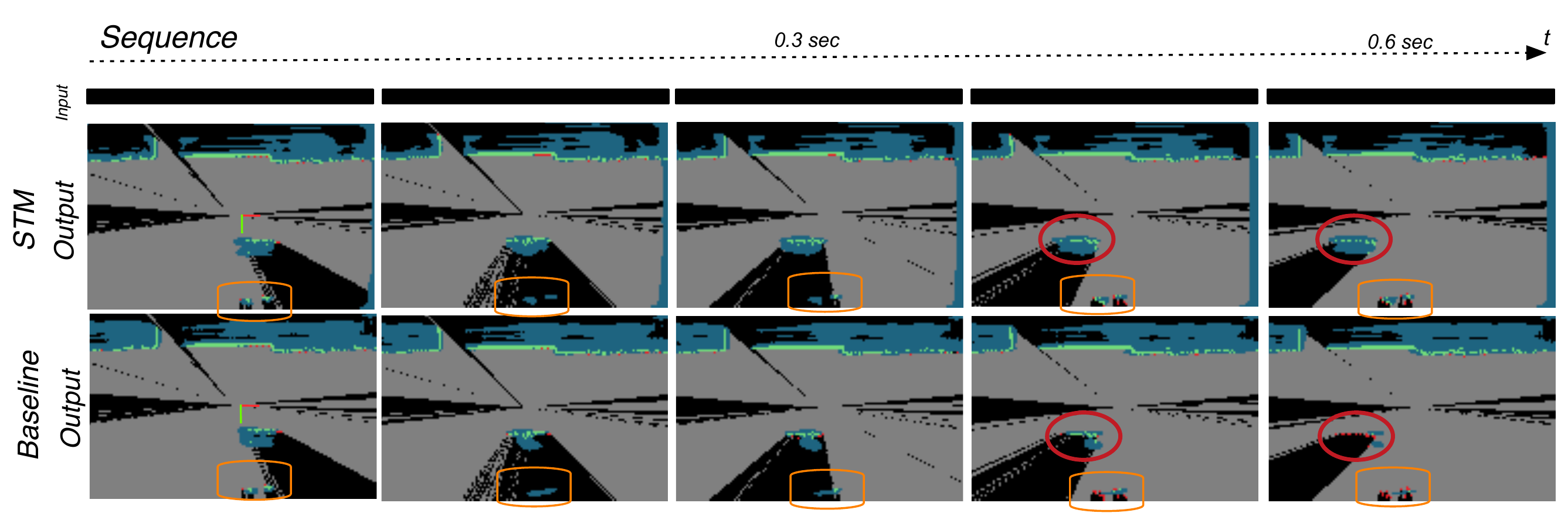}
\caption{Left to Right: Output sequence (0.6 seconds) of dynamic and static object tracking through occlusion as predicted by the Spatial Transformer Model (Top) and the Baseline Model (Bottom). We show the colour-coded comparison to the visible ground truth, with false negatives in red, false positives in blue, and true positives in green. Where the Spatial Transformer network consistently accurately tracks two pedestrians (orange rectangle) occluded by a moving vehicle (red), the Baseline model fails to capture the dynamics of both the vehicle and the pedestrians. This is particularly visible on the last frame, with most of the area around the actual location of these objects showing false negative occupancy prediction for the Baseline model.}
\label{fig:ST2}
\end{figure*}

\paragraph{Quantitative Results}
Figure~\ref{fig:F1moving} represents the F1 measure as computed with STM and with the baseline DT.
The baseline DT is identical to STM with the exception that no egomotion is taken into account. In other words, no additional mask is applied to the cost computation and backpropagation, and hidden states are not forward transformed into the next sensor frame.

With no egomotion information, one might expect the F1 measure for the baseline to be very poor. Surprisingly however, it does very well as illustrated in Figure~\ref{fig:F1moving}. We suggest two explanations for this. Firstly, we posit that this is due to the dataset being relatively benign in terms of vehicle motion patterns. As most of the driving occurs down straight roads and at relatively constant velocity ($\sim20$ mph) the baseline may have learned a constant velocity model that could be used to correct the hidden state update.
Secondly, the F1 measure here may be less informative, as the dataset is dominated by static objects such as walls and buildings. 
As such, a large fraction of the F1 score can be attained by learning the vehicle's motion and merely capturing static scenes.
Nonetheless, the STM offers a clear improvement over the baseline DT in all future frames.


\subsubsection{Qualitative Results}
To qualitatively evaluate the model, we show a selection of sequences where the model does well, and where it does more poorly. As the dataset does not contain as many occlusions and for as long as the static set due to the setting of a vehicle driving through an urban environment, we look at the ability of the network to  predict what happens in occlusion by maintaining the blacking out of the input every $5$ frames at test time.

Figure~\ref{fig:ST1} shows a compelling example of STM accurately tracking both dynamic and static objects through occlusion. 
In particular, the model accurately predicts the position of two static objects of different sizes when future frames are blacked out, and is also able to maintain the tracks when both objects are fully occluded.

Figure~\ref{fig:ST2} illustrates how STM accurately predicts the tracks of both a moving vehicle (red circle) and two occluded pedestrians (orange rectangle) whereas the baseline model fails. For the latter, the predicted track of the vehicle gradually shifts from the ground truth until complete failure in frame 5 (false negative area), and it fails to separate and track the pedestrians through occlusion by the vehicle.

\section{Conclusion}
\label{sec:conclusion}
In this paper, we have proposed an approach to perform object tracking for a mobile robot travelling in crowded urban environments, building on the previously proposed \emph{DeepTracking} framework (\cite{OndruskaAAAI2016, ondruska2016end}).
Crucially, unlike classical techniques which employ a multi-stage pipeline, this approach is learned end-to-end with limited architectural choices.
By employing a \emph{Spatial Transformer} module, the model is able to exploit noisy estimates of visual egomotion as a proxy for true vehicle motion.
Experimental results demonstrate that our method performs favourably to \emph{DeepTracking} in terms of accurately predicting future states, and show that the model can capture the location and motion of cars, pedestrians, cyclists, and buses, even when in complete occlusion.
\vspace{0.5cm}

Future work will look to estimate ego-motion, explore modalities such as radar, and extend the approach to 3D.

\section*{Acknowledgement}
The authors would like to gratefully acknowledge support of this work by the UK Engineering and Physical Sciences Research Council (EPSRC) Doctoral Training Programme (DTP) and Programme Grant DFR-01420, as well as the Advanced Research Computing services at the University of Oxford for providing access to their computing cluster.

\bibliographystyle{IEEEtran}
\bibliography{references}

\end{document}